# Explainable Deep Learning for Pediatric Pneumonia Detection in Chest X-Ray Images

Adil O. Khadidos[1], †, Aziida Nanyonga[2]*, †, Alaa O. Khadidos[3,4], Olfat M. Mirza[5,] Mustafa Tahsin Yilmaz[6,4]

[1]Department of Information Technology, Faculty of Computing and Information Technology, King Abdulaziz University, Jeddah, Saudi Arabia

[2]School of Science and Technology, Equator University of Science and Technology, Masaka, Uganda; dr.nanyonga@equsat.ac.ug

[3] Department of Information Systems, Faculty of Computing and Information Technology, King Abdulaziz University, Jeddah, Saudi Arabia

[4] Centre of Research Excellence in Artificial Intelligence and Data Science (AIADS), King Abdulaziz University, Jeddah, Saudi Arabia

[5] Department of Computer Science, College of Computers and Information Systems, Umm Al-Qura University, Makkah, Saudi Arabia

[6] Department of Industrial Engineering, Faculty of Engineering, King Abdulaziz University, Jeddah, Saudi Arabia

† These authors contributed equally to this work.

\* Correspondence: dr.nanyonga@equsat.ac.ug

**Abstract**

**Background**: Pneumonia remains a major cause of illness and death among children worldwide, highlighting the need for accurate and efficient diagnostic support tools. Deep learning has shown strong potential in medical image analysis, particularly for chest X-ray interpretation. This study aims to compare two state-of-the-art convolutional neural network architectures for automated pediatric pneumonia detection.

**Methods**: A publicly available dataset of 5,863 pediatric chest X-ray images was used. All images were preprocessed through normalization, resizing, and augmentation to improve model generalization. DenseNet121 and EfficientNet-B0 were fine-tuned using pretrained ImageNet weights under identical training conditions. Model performance was evaluated using accuracy, F1-score, Matthews Correlation Coefficient (MCC), and recall. Explainability was incorporated using Gradient-weighted Class Activation Mapping (Grad-CAM) and Local Interpretable Model-agnostic Explanations (LIME) to visualize regions contributing to model decisions.

**Results**: EfficientNet-B0 achieved superior classification performance, with an accuracy of 84.6%, F1-score of 0.8899, and MCC of 0.6849. DenseNet121 obtained 79.7% accuracy, an F1-score of 0.8597, and an MCC of 0.5852. Both models achieved high recall values above 0.99, indicating strong sensitivity to pneumonia detection. Grad-CAM and LIME visualizations confirmed that both models consistently highlighted clinically relevant lung regions, enhancing interpretability and supporting the reliability of predictions.

**Conclusions**: EfficientNet-B0 demonstrated a more balanced and computationally efficient performance compared to DenseNet121, making it a strong candidate for real-world clinical deployment. The integration of explainability techniques ensures transparency in model decision-making, contributing to the development of trustworthy AI-assisted diagnostic tools for pediatric pneumonia detection.

**Keywords: D**eep Learning; chest X-ray; pneumonia detection; EfficientNet-B0; DenseNet121; explainable AI; Grad-CAM; LIME; pediatric radiology

## 1. Background

Pneumonia remains one of the leading causes of morbidity and mortality among children under five years of age, accounting for approximately 14% of all global child deaths and responsible for over 800,000 deaths annually (WHO, 2021)[2]. Despite major public health interventions, the burden of pneumonia continues to be disproportionately high in low- and middle-income countries (LMICs), where healthcare infrastructure, diagnostic resources, and trained radiological expertise are often limited [3, 4]. The World Health Organization (WHO) and UNICEF have repeatedly emphasized the urgent need for scalable diagnostic support systems to improve early detection and management of pneumonia in these settings [5].

Chest radiography (CXR) is regarded as the clinical gold standard for diagnosing pneumonia because it allows visualization of lung opacities and infiltrates [6]. However, radiograph interpretation is inherently subjective and dependent on radiologist experience, leading to significant inter-observer variability, particularly in borderline or atypical cases [7]. Moreover, in resource-limited settings, the scarcity of radiologists and high workload contribute to diagnostic delays and inconsistencies [8]. Consequently, the development of automated, accurate, and interpretable computer-aided diagnostic (CAD) systems has become a priority for enhancing diagnostic capacity and supporting clinical decision-making in such environments.

Recent advances in artificial intelligence (AI), intense learning, have revolutionized medical image analysis, enabling automated feature extraction and classification without manual preprocessing [9-11]. Convolutional neural networks (CNNs) have achieved state-of-the-art performance across diverse radiological applications, including the detection of tuberculosis, pneumonia, lung nodules, and COVID-19 from chest X-rays [12-14]. Transfer learning has further accelerated this progress by allowing pretrained models, initially trained on large datasets such as ImageNet, to be fine-tuned for domain-specific medical tasks, achieving high diagnostic accuracy even with limited annotated medical data [15].

Among modern CNN architectures, DenseNet [16] and EfficientNet [17] stand out for their balance of accuracy and computational efficiency. DenseNet promotes feature reuse through dense connectivity, mitigating vanishing gradients and improving parameter efficiency, whereas EfficientNet introduces compound scaling to optimize model depth, width, and resolution. These architectures have demonstrated outstanding results in medical imaging tasks such as pneumonia and COVID-19 detection [18-20]. However, despite impressive performance, these models largely operate as "black boxes," offering little insight into their decision processes. The lack of interpretability poses a major barrier to clinical adoption, as physicians and regulators demand transparency to ensure algorithmic reliability, safety, and accountability [21-23].

To address this limitation, explainable artificial intelligence (XAI) has emerged as a crucial paradigm for developing transparent and trustworthy AI systems in healthcare [24]. Post-hoc interpretability methods, such as Gradient-weighted Class Activation Mapping (Grad-CAM) [25] and Local Interpretable Model-Agnostic Explanations (LIME) [26] enable visualization of model attention regions or local decision boundaries, helping clinicians verify whether the model's focus corresponds to clinically meaningful patterns such as lung consolidations or infiltrates. These techniques foster human–AI collaboration by providing visual explanations that align computational reasoning with medical interpretation, ultimately enhancing clinical trust and decision support [27, 28].

In this study, we propose an explainable deep learning framework for pediatric pneumonia detection using chest X-rays, leveraging transfer learning with EfficientNet-B0 and DenseNet121 architectures. Our approach integrates Grad-CAM and LIME to generate visual, interpretable explanations of model predictions, allowing both quantitative and qualitative evaluation of diagnostic and interpretability performance. Beyond achieving high accuracy, this work emphasizes model transparency and clinical relevance as essential components for responsible AI integration in medical imaging workflows.

The key contributions of this work are summarized as follows:

1. Comparative evaluation of two state-of-the-art convolutional neural network (CNN) architectures, EfficientNet-B0 and DenseNet121, fine-tuned via transfer learning for pediatric pneumonia detection from chest X-ray images.
2. Integration of explainable artificial intelligence (XAI) techniques, specifically Gradient-weighted Class Activation Mapping (Grad-CAM) and Local Interpretable

Model-Agnostic Explanations (LIME), to enhance model interpretability and transparency.

3. Comprehensive quantitative and qualitative analysis of model performance across multiple evaluation metrics, including accuracy, precision, recall, F1-score, ROC-AUC, MCC, and Cohen's Kappa, combined with visual interpretability assessment.

4. Critical examination of clinical relevance, demonstrating how explainability methods can bridge the trust gap between AI-driven predictions and radiological reasoning, thus supporting potential real-world integration in diagnostic workflows.

The remainder of this paper is organized as follows: Section II reviews related work in AI-driven pneumonia detection and explainability methods. Section III details the datasets, preprocessing, models, and interpretability approaches employed. Section IV presents experimental results along with a comprehensive discussion. Finally, Section V concludes the paper and outlines future directions for explainable AI in medical imaging.

## 2. Related work

*2.1 Deep Learning in Medical Imaging*

Deep learning has revolutionized medical image analysis, enabling breakthroughs across tasks such as tumour segmentation, diabetic retinopathy screening, and thoracic disease classification. Convolutional Neural Networks (CNNs), in particular, have achieved state-of-the-art performance due to their ability to automatically extract hierarchical features from raw image data [9, 10].

Kermany et al. [13] pioneered the use of CNNs for pediatric pneumonia detection using chest X-rays, achieving diagnostic performance comparable to expert radiologists. This seminal work inspired a wave of studies leveraging transfer learning and advanced CNN architectures, such as DenseNet [16] and EfficientNet [17] to enhance diagnostic accuracy.

Recent developments have expanded beyond classical CNNs toward hybrid and transformer-based architectures to capture long-range dependencies and global context in chest X-ray interpretation. Oltu et al. [29] proposed a hybrid model combining DenseNet201 and Vision Transformers (ViT), outperforming several state-of-the-art baselines on multiclass classification tasks. Similarly, Hasan et al. [30] Conducted a comprehensive review of deep learning models for pneumonia detection, highlighting how attention-augmented CNNs and multi-stage feature extractors improved performance on large chest X-ray datasets.

In another study, Siddiqi [31] emphasized the growing role of explainable and efficient CNN models, such as EfficientNet-B0, which achieved high accuracy while maintaining computational efficiency, critical for deployment in low-resource settings. Collectively, these studies underscore deep learning's transformative potential in automating radiological workflows.

*2.2 Pneumonia Detection in Chest X-Rays*

Publicly available datasets such as NIH ChestX-ray14 [32], RSNA Pneumonia Challenge [33] and the pediatric chest X-ray dataset [1] have driven advancements in pneumonia classification research.

Jiang et al. [34] demonstrated that a VGG16-based CNN with background-aware preprocessing achieved 95.6% accuracy in pneumonia detection. Sheu et al. [35] further enhanced interpretability by integrating human-in-the-loop explanations within a transfer learning framework, achieving accuracy above 93%.

Beyond conventional CNNs, researchers have also developed custom and hybrid architectures. For instance, Shah et al. [36] proposed a data-driven framework combining CNNs with explainable AI techniques for chest X-ray classification, achieving high diagnostic accuracy while emphasizing model interpretability. Similarly, Rabbah et al. [37] integrated Inception-v3 with dense layers and Integrated Gradients (IG) for interpretable pediatric pneumonia diagnosis.

Despite these improvements, many studies still prioritize predictive accuracy over explainability, limiting their adoption in clinical environments. Radiologists and clinicians require interpretable insights such as heatmaps or feature attribution to understand model decisions and validate diagnostic relevance.

*2.3 Explainable AI in Medical Imaging*

Deep learning models often operate as "black boxes," hindering their clinical translation despite their diagnostic power. To bridge this gap, Explainable Artificial Intelligence (XAI) methods have emerged as indispensable tools for promoting model transparency and clinician trust. Among these, Gradient-weighted Class Activation Mapping (Grad-CAM) [25] and Local Interpretable Model-Agnostic Explanations (LIME) [26] are the most widely used. These approaches generate visual heatmaps and feature-level explanations that help correlate model attention with pathologically relevant regions in chest X-rays.

Recent research highlights increasing efforts toward more comprehensive explainability frameworks. Houssein and Gamal [38] provided an extensive review of XAI in medical imaging, emphasizing how hybrid visualization techniques improve diagnostic reliability. Cervantes & Chan demonstrated that incorporating LIME explanations significantly improved clinician confidence in CNN predictions [39]. Similarly, Erukude et al. [40], combined deep learning and explainable AI for pneumonia and brain tumor detection, producing interpretable predictions that aligned closely with expert annotations. These studies collectively affirm that integrating XAI into medical imaging models not only enhances transparency but also supports ethical and regulatory compliance.

*2.4 Research Gap*

While prior research confirms the potential of deep learning and XAI in medical imaging, several gaps persist. Most existing studies either emphasize predictive accuracy (e.g., hybrid CNN–ViT models, attention-based networks) or focus exclusively on interpretability (e.g., Grad-CAM, LIME, Integrated Gradients) without jointly evaluating both. Moreover, the majority of XAI research has been conducted on adult datasets such as NIH, CheXpert, and MIMIC-CXR, leaving pediatric imaging underexplored.

Our work addresses this gap by (i) evaluating two high-performing CNN architectures (EfficientNet-B0 and DenseNet121) using transfer learning, (ii) systematically comparing Grad-CAM and LIME explanations, and (iii) emphasizing both quantitative accuracy and qualitative interpretability. This dual-focus approach bridges the divide between algorithmic performance and clinical transparency, contributing to more explainable and trustworthy pediatric pneumonia diagnostics.

Table 1. Summary of Related Studies in Pneumonia Detection with Deep Learning and XAI

| Study | Dataset(s) | Model(s) | Explainability | Key Findings | Gap Addressed |
|---|---|---|---|---|---|
| Kermany et al. **[13]** | Pediatric CXR | Custom CNN | None | Accuracy comparable to radiologists | No interpretability |
| Rajpurkar et al. [12] | NIH ChestX-ray14 | CheXNet (DenseNet121) | None | High AUC for pneumonia detection | Focused on performance |
| Muchina et al. [41] | Private CXR | Multiple CNNs | LIME | Improved local interpretability | LIME only |
| Cervantes & Chan [39] | Public/Private (COVID/pneumonia) | Multiple CNNs | LIME | Improved local interpretability | Focused on LIME only |
| Sheu et al. **[35]** | NIH + VinDr | DCNN + TL | XAI-ICP | 93% accuracy, interpretable | No Grad-CAM |
| Jiang et al. **[34]** | NIH CXR | VGG16 | Grad-CAM | Accuracy improved with preprocessing | Single model |
| Colin & Surantha, [42] | NIH CXR | ResNet50 | LRP, CAMs | Best interpretability-performance trade-off | No DenseNet/EfficientNet |
| **Our Study** | **Pediatric CXR** | **EfficientNet-B0, DenseNet121** | **Grad-CAM, LIME** | **Joint interpretability + many performance metrics** | **Bridges performance and transparency** |

## 3. Methods

The methodology adopted in this study was designed to ensure a rigorous, fair, and clinically meaningful evaluation of deep learning models for pediatric chest X-ray classification. In particular, we compare DenseNet121 and EfficientNet-B0, both state-of-the-art CNN architectures, to assess their effectiveness in distinguishing normal chest radiographs from those with pneumonia. Each step of the process, from dataset selection to explainability, was carefully curated to enhance both scientific rigour and clinical relevance.

Figure 1 provides an overview of the methodological framework employed in this study, illustrating the sequential stages involved in developing and evaluating the proposed models. The workflow begins with the acquisition and preprocessing of the pediatric chest X-ray dataset, followed by model design and fine-tuning of DenseNet121 and EfficientNet-B0 architectures. Both models were trained and validated under identical experimental configurations to ensure a fair comparison. The final stage of the pipeline includes performance evaluation using comprehensive quantitative metrics and qualitative explainability techniques, thereby providing both computational robustness and clinical interpretability. This structured workflow ensures methodological consistency, reproducibility, and transparency throughout the research process.

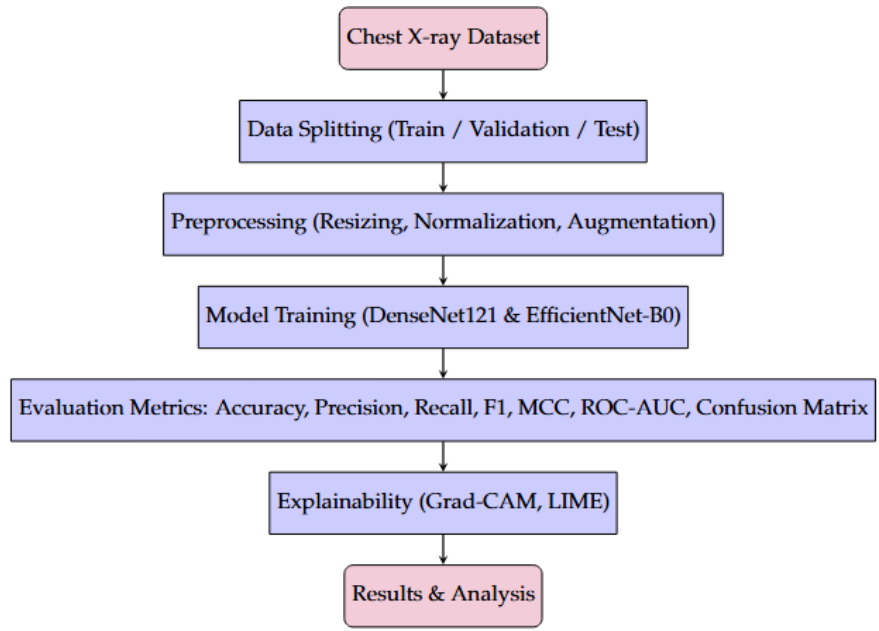

Figure 1 shows the methodological overview of this research.

*3.1 Dataset*

This study employed the publicly available pediatric chest X-ray dataset introduced by Kermany et al. [1]. The dataset consists of 5,863 anterior–posterior (AP) chest radiographs from children between the ages of one and five years. Each image was carefully labelled into one of two diagnostic categories: *Normal* or *Pneumonia*. To ensure annotation accuracy, all labels were initially assigned by two expert physicians, with disagreements adjudicated by a third radiologist. This consensus-driven process strengthens the reliability of the dataset, reducing annotation bias and providing a solid foundation for training medical imaging models.

The dataset was partitioned into training, validation, and testing subsets as provided by the original authors. Importantly, it is distributed under a Creative Commons Attribution 4.0 (CC BY 4.0) license, ensuring its ethical and transparent use in academic research.

*3.2 Image Preprocessing*

Medical imaging data typically exhibit heterogeneity in terms of resolution, illumination, and orientation, which may hinder the direct application of deep learning models.

Consequently, a preprocessing pipeline was implemented to standardize the input data and improve model generalization.

All radiographs were resized to 224 × 224 pixels, corresponding to the input resolution required by both DenseNet121 and EfficientNet-B0, which were originally trained on ImageNet [43]. Pixel intensities were normalized to fall within the range [0, 1]. To further align the images with the pretrained weights, mean and standard deviation normalization was performed using ImageNet statistics, with μ = [0.485, 0.456, 0.406] and σ = [0.229, 0.224, 0.225] [44].

Given the relatively limited size of the dataset, data augmentation played a critical role in improving robustness and mitigating overfitting. Augmentations included random horizontal flips (p = 0.5) to simulate mirrored radiographs, rotations within ±15° to account for variations in patient positioning, and random zooming and scaling in the range 0.9–1.1 to mimic differences in imaging distance. Brightness adjustments were also applied to replicate variations in radiographic intensity. These transformations generated diverse yet realistic training samples, thereby enabling the models to learn invariant and clinically relevant features [45].

*3.3 Deep Learning Architectures*

Two convolutional neural network architectures were selected for this study: DenseNet121 and EfficientNet-B0. These models were chosen based on their proven performance in medical imaging and their complementary design philosophies.

DenseNet121 employs a densely connected architecture in which each layer receives inputs from all preceding layers [16]. This dense connectivity promotes feature reuse, enhances gradient flow, and significantly mitigates the vanishing gradient problem. DenseNet121 consists of 121 layers grouped into four dense blocks, with transition layers composed of batch normalization, a 1×1 convolution, and 2×2 average pooling operations. The network is highly parameter-efficient, achieving strong performance with fewer weights compared to traditional CNNs. For this study, DenseNet121 was initialized with pretrained ImageNet weights and fine-tuned for binary classification by replacing the final fully connected layer with a two-class classifier.

EfficientNet-B0, by contrast, was designed using a compound scaling method that systematically balances network depth, width, and resolution [17]. Its architecture builds upon Mobile Inverted Bottleneck Convolution (MBConv) layers, enhanced with squeeze-and-excitation blocks for channel-wise attention [46]. EfficientNet-B0 is notable for achieving a superior accuracy-to-efficiency trade-off compared to conventional CNNs, making it computationally lightweight yet powerful. Like DenseNet121, EfficientNet-B0 was initialized with ImageNet-pretrained weights, and its classifier head was modified for binary classification. Together, the inclusion of both DenseNet121 and EfficientNet-B0 provides an opportunity to compare a densely connected architecture against a scaled and efficiency optimized model.

*3.4 Training Setup*

To ensure a fair and reproducible comparison, both models were trained under identical experimental conditions. The Adam optimizer [47], was selected with an initial learning rate of $1 \times 10^{-4}$ due to its adaptive learning rate capability, which balances convergence speed and stability. The Binary Cross-Entropy Loss function was employed to penalize misclassifications, as it is particularly suitable for two-class problems. Training was conducted using a batch size of 32, which balances computational efficiency with stable gradient updates.

Both networks were trained for a maximum of 10 epochs, although early stopping was applied to prevent overfitting. Specifically, training was terminated if validation loss failed to improve for three consecutive epochs. A ReduceLROnPlateau learning rate scheduler was also employed to decrease the learning rate dynamically when the validation loss plateaued, enabling finer convergence in later epochs. All experiments were executed on an NVIDIA GPU within a CUDA 11.x environment, ensuring efficient parallel computation.

*3.5 Evaluation Metrics*

Model performance was evaluated using a comprehensive set of metrics, as reliance on accuracy alone can be misleading in medical diagnosis tasks [48]. Accuracy measured the overall proportion of correctly classified images, while precision quantified the reliability of positive pneumonia predictions, an important factor for avoiding false alarms. Recall, also known as sensitivity, assessed the ability of the models to correctly identify pneumonia cases, which is critical in reducing the likelihood of missed diagnoses. The F1-score, representing the harmonic mean of precision and recall, was included as a balanced performance indicator.

In addition, confusion matrices were generated to provide a visual representation of classification outcomes, revealing the distribution of true positives, true negatives, false positives, and false negatives. The Receiver Operating Characteristic (ROC) curve and the corresponding Area Under the Curve (AUC) were employed to evaluate the discriminative capability of the models across different decision thresholds [49]. The Matthews Correlation Coefficient (MCC) was also reported, as it offers a balanced evaluation even in the presence of imbalanced datasets [50]. Collectively, these metrics provide a thorough assessment of the models from both clinical and computational perspectives.

*3.6 Explainability Methods*

To enhance the interpretability and clinical trustworthiness of the deep learning predictions, two complementary model explainability techniques were integrated into the proposed framework: Gradient-weighted Class Activation Mapping (Grad-CAM) and Local Interpretable Model-agnostic Explanations (LIME).

Grad-CAM highlights class-discriminative regions within chest radiographs by leveraging the gradients of the target class flowing into the final convolutional layers of the CNN [25]. This produces visual heatmaps that localize regions contributing most strongly to the model's prediction, thereby allowing clinicians to assess whether the model focuses on relevant pulmonary structures. In the context of pneumonia detection, Grad-CAM provides an intuitive visualization of lung regions associated with inflammatory or opacified areas, facilitating a better understanding of how the model distinguishes between normal and abnormal images.

Complementarily, LIME was applied to provide local, instance-level interpretability by perturbing input images into superpixel segments and evaluating how changes in each segment affect the model's output [26]. This generates a simplified, interpretable approximation of the model's decision boundary for individual predictions. By identifying the contribution of each superpixel to the final classification, LIME enables verification of whether the model's attention corresponds to clinically meaningful regions such as the lung fields rather than irrelevant artifacts.

Together, Grad-CAM and LIME offer both global (model-level) and local (instance-level) interpretability. Their integration ensures that the proposed CNN classifiers are not only accurate but also transparent and clinically explainable, addressing a critical challenge in the adoption of artificial intelligence systems in healthcare diagnostics. Representative visualizations generated using Grad-CAM and LIME are presented in Section 4 (Results), illustrating how both methods contribute to the interpretability of the model predictions.

## 4. Results

*4.1. Model Performance*

Both DenseNet121 and EfficientNet-B0 were trained on the pediatric chest X-ray dataset for binary classification of Normal versus Pneumonia. Their performance was assessed using multiple metrics to capture different aspects of diagnostic reliability, including accuracy, precision, recall, F1-score, ROC-AUC, Matthews Correlation Coefficient (MCC), Cohen's Kappa, and Brier Score. As shown in Table 1, EfficientNet-B0 outperformed DenseNet121 across most evaluation metrics, achieving higher Accuracy (0.8462 vs. 0.7965), Precision (0.8050 vs. 0.7553), F1-score (0.8899 vs. 0.8597), MCC (0.6849 vs. 0.5852), and Cohen's Kappa (0.6438 vs. 0.5139). The lower Brier Score of 0.1302 for EfficientNet-B0 compared to 0.1602 for DenseNet121 further confirms its superior calibration and predictive confidence.

Table 1 summarizes the comparative performance of the two models.

| Metric | DenseNet121 | EfficientNet-B0 |
| --- | --- | --- |
| Accuracy | 0.7965 | 0.8462 |
| Precision | 0.7553 | 0.8050 |
| Recall | 0.9974 | 0.9949 |
| F1-score | 0.8597 | 0.8899 |
| ROC-AUC | 0.9755 | 0.9652 |
| MCC | 0.5852 | 0.6849 |
| Cohen's Kappa | 0.5139 | 0.6438 |
| Brier Score | 0.1602 | 0.1302 |

While DenseNet121 exhibited slightly higher Recall (0.9974 vs. 0.9949) and ROC-AUC (0.9755 vs. 0.9652), these differences are minimal. Both models demonstrate strong discriminative ability (AUC > 0.96), meaning they can correctly rank pneumonia-positive cases with >96% probability. These findings align with previous research emphasizing the effectiveness of transfer learning in medical imaging [13, 51].

The advantage of EfficientNet-B0 is attributed to its compound scaling strategy, which balances network depth, width, and resolution to optimize accuracy efficiency trade-offs [17]. This results in more stable and calibrated predictions, confirming its suitability for lightweight clinical AI applications

4.1.1. ROC and Precision–Recall Curves

As shown in Figure 2, the near-identical AUCs (0.965–0.975) reaffirm both models' strong discriminative capacity. However, the slightly higher precision and F1-score for EfficientNet-B0 indicate more consistent positive predictions across varying thresholds.

As illustrated in Figure 3, the Precision, Recall and F1-Score bar graph confirms that EfficientNet-B0 maintains better balance under class-imbalance conditions, a critical property in medical datasets where pneumonia-positive cases often dominate [52]. This robustness reduces the likelihood of false alarms without sacrificing sensitivity.

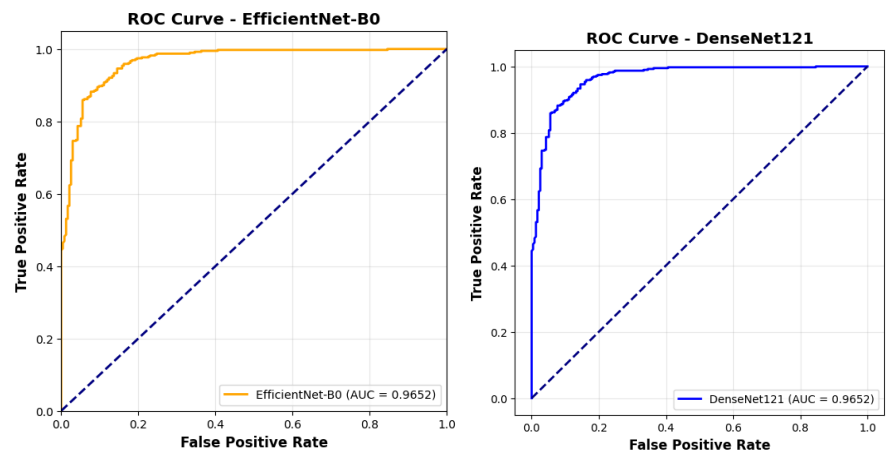

Figure 2 presents the ROC curves for both models.

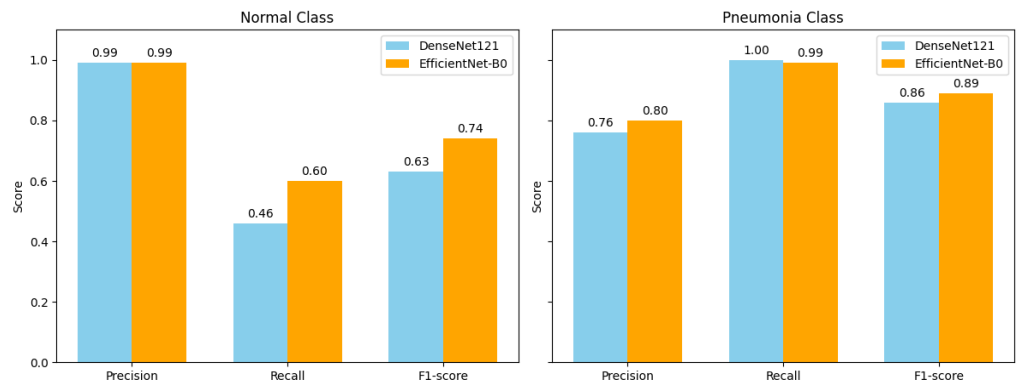

Figure 3 shows the Precision, Recall and F1-score of both models.

*4.2. Confusion Matrix Analysis*

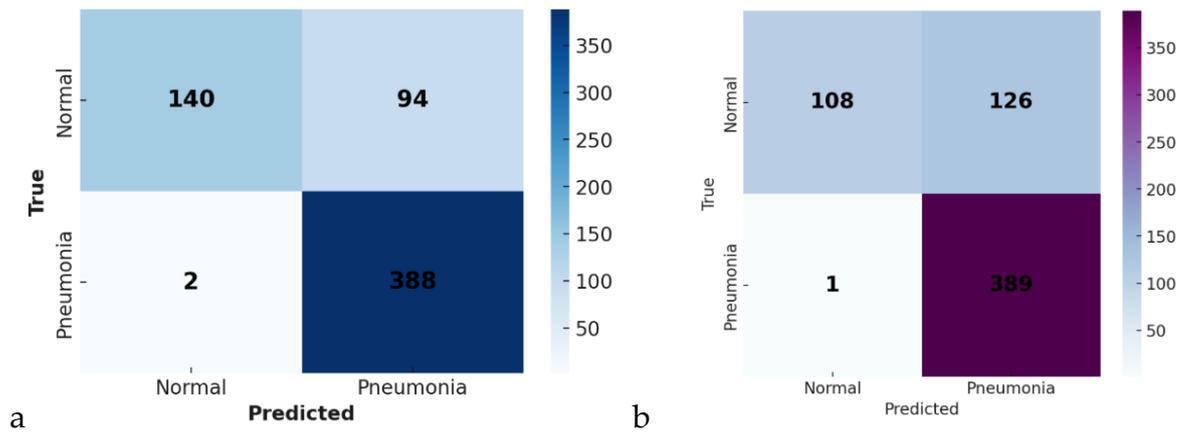

Figure 4 shows the confusion matrix (a) EfficientNet-B0 and (b) DenseNet121

The confusion matrices for both models (Figure 4a–b) provide further insight into classification behavior. EfficientNet-B0 correctly identified 388 out of 390 pneumonia cases and 140 out of 234 normal cases, while DenseNet121 detected 389 out of 390 pneumonia cases but misclassified a larger number of normal cases (126 false positives).

This trend indicates that DenseNet121 is slightly more sensitive but less specific, prioritizing the detection of all pneumonia cases even at the expense of more false alarms. In contrast, EfficientNet-B0 achieves a better balance between sensitivity and specificity, minimizing overdiagnosis while maintaining near-perfect recall. From a clinical standpoint, such balance is crucial: while missed pneumonia cases can be life-threatening, excessive false positives can lead to unnecessary treatments or hospital referrals.

*4.3. Explainability: Grad-CAM and LIME Analysis*

To ensure interpretability, Grad-CAM and LIME visualizations were generated for representative test cases. Figures 5 and 6 show Grad-CAM heatmaps for Normal and Pneumonia images, respectively. The highlighted activation regions correspond well with clinically relevant lung areas, especially the lower lobes and perihilar regions, common sites of pneumonia manifestation.

LIME explanations (Figures 7 and 8) further validate these findings by isolating superpixel regions that contribute most to the model's predictions. In correctly classified pneumonia cases, the red-highlighted areas in both Grad-CAM and LIME outputs align with radiopaque infiltrates observable on the chest X-rays. For normal cases, activations were dispersed across non-pathological areas, suggesting the models learned to focus on relevant patterns rather than noise or artifacts. Figure 9 demonstrates Grad-CAM intensity, where red regions correspond to high activation (model confidence of 0.87 for a true pneumonia image). Figure 10 illustrates representative chest X-ray images from the dataset, depicting examples of both Normal and Pneumonia cases used for model training and evaluation.

Such interpretability aligns with recent trends emphasizing explainable AI in radiology [35, 42]. Compared to earlier CNN models that acted as "black boxes," our dual-explainability approach demonstrates that both EfficientNet-B0 and DenseNet121 can yield clinically meaningful visual rationales. This helps bridge the gap between algorithmic performance and clinical trustworthiness.

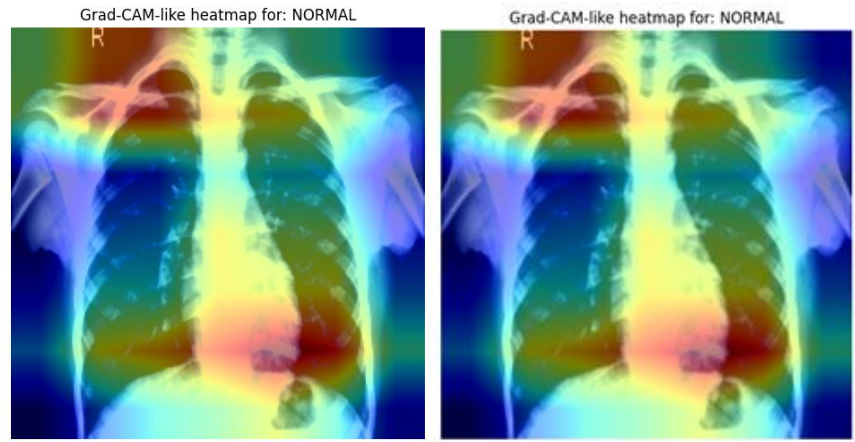

**Figure 5.** Shows Grad-CAM heatmaps for Normal images.

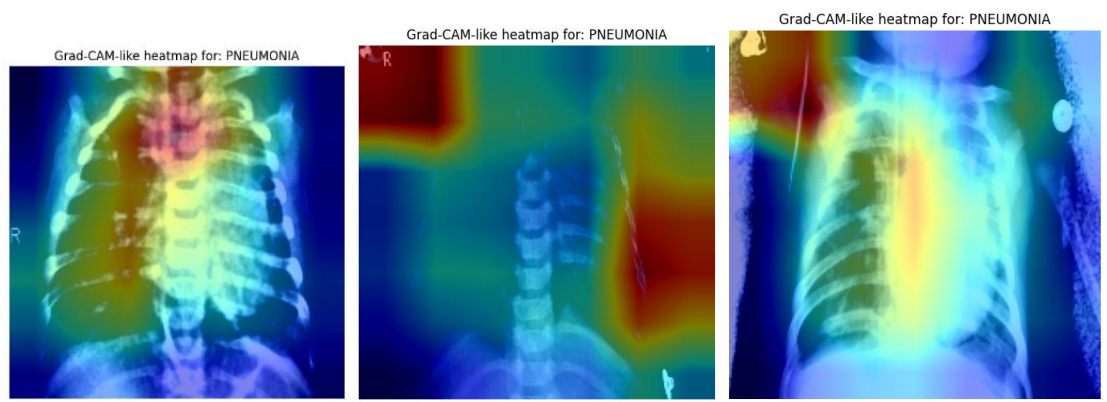

**Figure 6.** Shows Grad-CAM heatmaps for Pneumonia images.

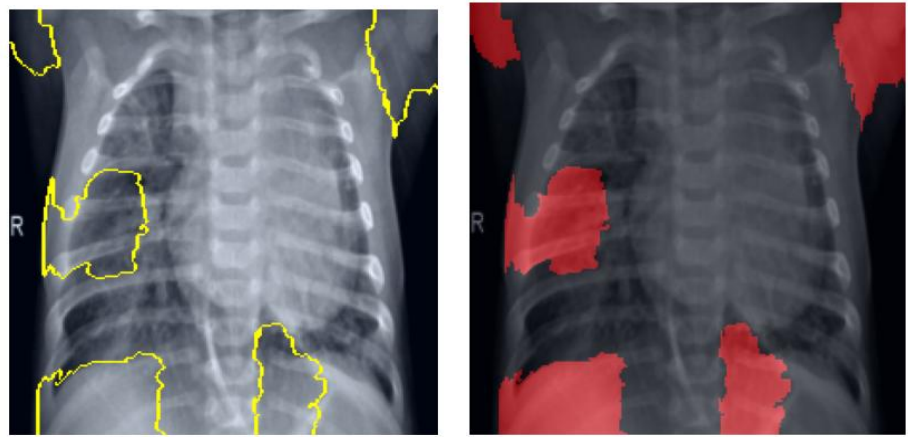

**Figure 7.** Shows LIME explanation for Pneumonia images.

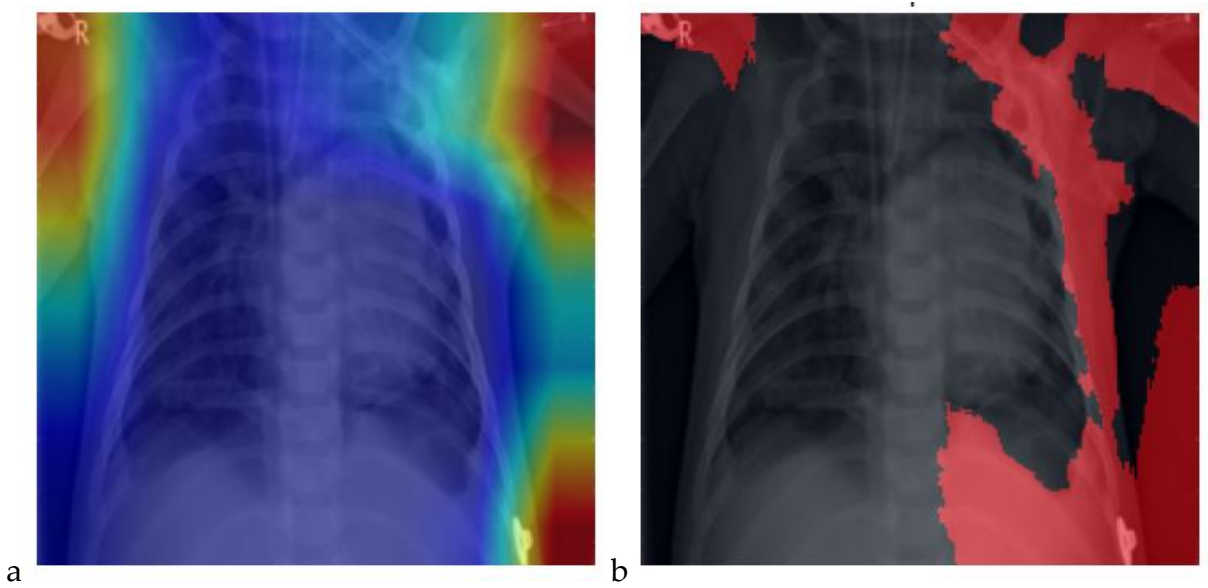

**Figure 8**. Shows (a) Grad-CAM Heatmap, while (b) shows Lime explanation.

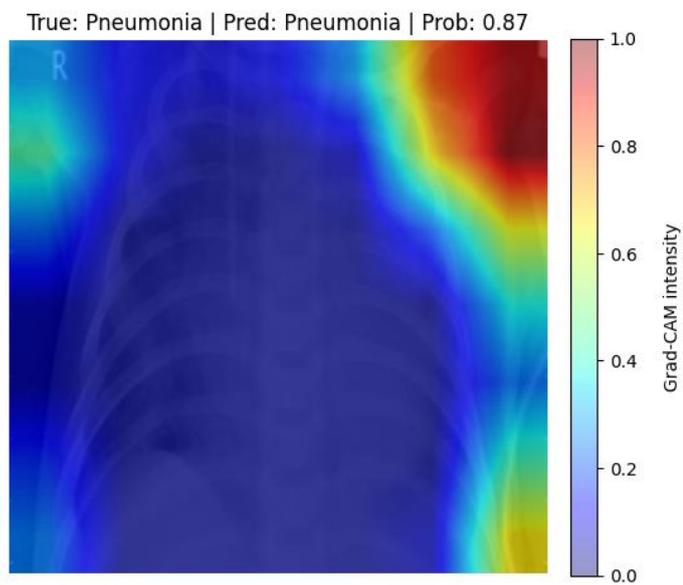

Figure 9. Shows the Grad-CAM intensity.

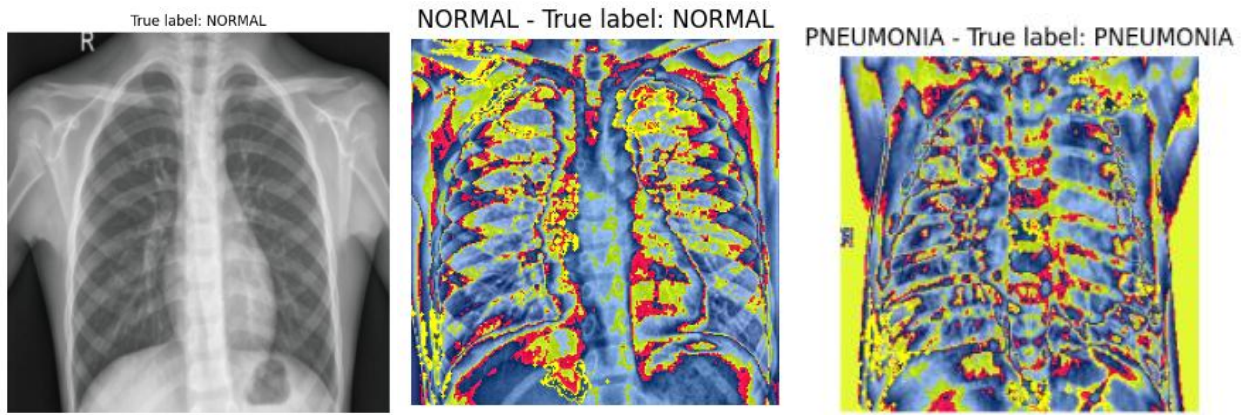

**Figure 10**. Shows the image of the true label.

## 5. Discussion

The observed results reinforce and extend findings from earlier literature. Similar to Kermany et al. [13] both models achieve diagnostic performance comparable to radiologists. However, by integrating Grad-CAM and LIME, this work enhances the transparency aspect often overlooked in prior studies [12, 53].

EfficientNet-B0's superior generalization and lower error rates mirror those of the hybrid and attention-based models [29, 51], which leveraged architectural innovations to improve efficiency and context understanding. Meanwhile, DenseNet121's higher recall parallels earlier observations by Rajpurkar et al. [12] where densely connected networks achieved robust feature reuse, making them highly sensitive to subtle pathological cues.

Overall, these findings demonstrate that EfficientNet-B0 offers a strong balance of diagnostic accuracy, computational efficiency, and interpretability, making it a compelling candidate for pediatric pneumonia detection in clinical decision-support systems.

*5.1. Ablation Study*

Despite strong results, several limitations must be acknowledged. The dataset originated from a single public source, which limits the diversity of imaging equipment, demographics, and disease patterns. Future studies should incorporate multi-centre and multi-ethnic datasets to improve generalization.

Additionally, the current study focused on binary classification (Normal vs. Pneumonia). Future work should extend this to multi-class differentiation (e.g., bacterial, viral, and COVID-19 pneumonia) and include auxiliary patient data such as age, symptoms, or laboratory findings for context-aware diagnosis.

While Grad-CAM and LIME enhance interpretability, they remain post-hoc methods whose visual explanations may vary with parameter settings. Future work could adopt intrinsically interpretable models or hybrid frameworks combining image and clinical data for explainable, robust decision support.

Moreover, optimizing inference time and resource efficiency will be critical for deployment in low-resource healthcare settings or mobile-based point-of-care systems. Exploring ensemble models that fuse DenseNet and EfficientNet architectures and leveraging federated learning for privacy-preserving training may further improve performance and trustworthiness.

*5.2. Limitations and Future Work*

Despite strong results, several limitations must be acknowledged. The dataset originated from a single public source, which limits the diversity of imaging equipment, demographics, and disease patterns. Future studies should incorporate multi-centre and multi-ethnic datasets to improve generalization.

Additionally, the current study focused on binary classification (Normal vs. Pneumonia). Future work should extend this to multi-class differentiation (e.g., bacterial, viral, and COVID-19 pneumonia) and include auxiliary patient data such as age, symptoms, or laboratory findings for context-aware diagnosis.

While Grad-CAM and LIME enhance interpretability, they remain post-hoc methods whose visual explanations may vary with parameter settings. Future work could adopt intrinsically interpretable models or hybrid frameworks combining image and clinical data for explainable, robust decision support.

## 6. Conclusions

This study presented a comparative evaluation of two state-of-the-art CNN models, DenseNet121 and EfficientNet-B0, for automated pediatric pneumonia detection from chest X-rays. Both models achieved high diagnostic accuracy, with EfficientNet-B0 demonstrating superior overall performance (Accuracy = 0.8462, F1 = 0.8899, MCC = 0.6849). DenseNet121 exhibited slightly higher Recall, underscoring its sensitivity advantage for identifying pneumonia cases.

The integration of Grad-CAM and LIME provided visual interpretability, ensuring that model predictions aligned with clinically relevant lung regions and supporting the potential of explainable AI in radiological decision-making.

While dataset and scope limitations exist, the findings highlight EfficientNet-B0's capability as a computationally efficient and interpretable solution for pneumonia screening. Future work will expand the framework to multi-class pneumonia types, incorporate clinical metadata, and optimize for real-time clinical deployment.

Therefore, this study contributes to the growing evidence that deep learning models, when carefully designed and explained, can augment radiologists' expertise, enhance diagnostic precision, and accelerate decision-making in pediatric care.

## Declarations

**Ethics approval and consent to participate:** Not applicable. This study used a publicly available, fully anonymized dataset (Kermany et al., 2018), which does not require institutional ethics approval.

**Consent for publication:** Not applicable. No identifiable human data are included in this study.

**Availability of data and materials:** The pediatric chest X-ray dataset analyzed in this study is publicly available: Kermany et al., (2018) Mendeley Data Repository https://data.mendeley.com/datasets/rscbjbr9sj/2. No new data were generated or collected for this study. The data were used strictly for academic and research purposes in accordance with the dataset's Creative Commons Attribution 4.0 (CC BY 4.0) license.

**Competing interests:** The authors declare that they have no competing interests.

**Funding:** This research was funded by the Deanship of Scientific Research (DSR) at King Abdulaziz University, Jeddah, Saudi Arabia, grant number IPP: 539-611-2025. The authors gratefully acknowledge DSR for financial and technical support.

**Author Contributions:** "Conceptualization, A.N. and A.K.; methodology, A.N.; software, A.N.; validation, A.K., O.M. and M.T.Y.; formal analysis, A.N.; investigation, A.N, A.K.; resources, A.K, A.N, M.T.Y.; data curation, A.N.; writing—original draft preparation, A.N, A.K. and M.T.Y.; writing—review and editing, M.T.Y.; visualization, A.K.; supervision, M.T.Y.; project administration, A.N.; funding acquisition, A.K. All authors have read and agreed to the published version of the manuscript.

## Abbreviations

The following abbreviations are used in this manuscript:

| Abbreviation | Full Term |
|---|---|
| AI | Artificial Intelligence |

| AP | Anterior–Posterior |
| --- | --- |
| AUC | Area Under the Curve |
| CAM | Class Activation Mapping |
| CNN | Convolutional Neural Network |
| CXR | Chest X-ray |
| DCNN | Deep Convolutional Neural Network |
| DL | Deep Learning |
| FC | Fully Connected |
| F1-score | Harmonic Mean of Precision and Recall |
| FN | False Negative |
| FP | False Positive |
| Grad-CAM | Gradient-weighted Class Activation Mapping |
| GPU | Graphics Processing Unit |
| IG | Integrated Gradients |
| LMIC | Low- and Middle-Income Countries |
| LIME | Local Interpretable Model-agnostic Explanations |
| LRP | Layer-wise Relevance Propagation |
| MCC | Matthews Correlation Coefficient |
| ReLU | Rectified Linear Unit |
| ROC | Receiver Operating Characteristic |
| SENet | Squeeze-and-Excitation Network |
| TP | True Positive |
| TN | True Negative |
| ViT | Vision Transformer |
| WHO | World Health Organization |
| XAI | Explainable Artificial Intelligence |